\documentclass[runningheads]{llncs}

\usepackage{eccv}

\usepackage{eccvabbrv}

\usepackage{graphicx}
\usepackage{booktabs}
\usepackage{makecell}

\usepackage[accsupp]{axessibility}  

\usepackage{hyperref}

\usepackage{orcidlink}

\usepackage{tikz}
\usetikzlibrary{arrows.meta,positioning,calc}
\usepackage{multirow}
\usepackage{float}

\begin{document}

\title{DINO-FIL: Multi-Layer Transformer Fusion for Generative Inpainting Forgery Localization} 

\titlerunning{DINO-FIL}

\author{Simon Freitter\inst{1}\orcidlink{0009-0006-4312-6203} \and
David Fischinger\inst{1}\orcidlink{0000-0002-0806-0468} \and
Silvia Poletti\inst{1}\orcidlink{0000-0002-5923-531X} \and
Marcel Hasenbalg\inst{1}\orcidlink{0000-0001-8012-8669} \and
Martin Boyer\inst{1}\orcidlink{0000-0002-0765-0829}}

\authorrunning{S.~Freitter et al.}

\institute{
AIT Austrian Institute of Technology GmbH, Giefinggasse 4,
1210 Vienna, Austria\\
\email{simon@freitter.com}
}
\maketitle

\begin{abstract}
  We propose DINO-FIL, a transformer-based framework for pixel-level localization of generative image forgeries. The method leverages multi-layer representations from a pretrained DINOv3 ViT-H+/16 encoder combined with a lightweight fusion decoder to capture both semantic structure and fine-grained texture inconsistencies introduced by image editing. By aggregating intermediate transformer features across depths, the model improves sensitivity to subtle, spatially localized artifacts that are often missed by standard single-layer approaches. We evaluate DINO-FIL in a cross-dataset setting, training on B-Free and testing on Beyond the Brush, CocoGlide, TGIF, and SAGI-D. The model achieves strong generalization, outperforming existing baselines on average and on four of five evaluated datasets, while maintaining a high true negative rate on pristine images.

\end{abstract}

\section{Introduction}
\label{sec:introduction}

Generative image editing has rapidly improved in visual quality and accessibility. Text-guided and mask-guided inpainting systems can now remove, replace, or regenerate selected image regions while preserving global scene coherence, making manipulated content increasingly difficult to recognize from visual inspection alone~\cite{giakoumoglou2025sagi}. Unlike conventional manipulations such as splicing or copy-move editing, generative inpainting can synthesize new content inside a user-specified mask and blend it into the surrounding context. This is particularly challenging in fully regenerated settings, where the edited region is synthesized jointly with its local visual context rather than inserted as an externally pasted region. As a result, traditional image forgery localization methods may not observe clear low-level discontinuities~\cite{mareen2024tgif,mareen2026tgif2}. Reliable localization of inpainted regions is therefore an important task for image forensics, media verification, and downstream trustworthiness analysis.

In this work, we address dense image inpainting localization. Given an RGB image, the goal is to predict a pixel-level binary mask that separates inpainted regions from authentic image content. Unlike image-level synthetic detection, dense inpainting localization targets a finer-grained, pixel-level task: identifying edited regions by detecting subtle statistical inconsistencies and visual artifacts, such as boundary discontinuities between authentic and synthesized content. In this setting, “edited” does not necessarily mean that the background pixels remain unchanged. Depending on the editing workflow, the background may either stay pixel-identical to the original image or be regenerated together with the modified region. In the latter case, pixel-level changes can occur across the entire image while preserving the background’s semantic content, potentially destroying the local forensic traces needed for inpainting localization, even though image-level synthetic detection may still succeed~\cite{mareen2026tgif2}.

Recent inpainting-focused benchmarks such as TGIF~\cite{mareen2024tgif}, Beyond the Brush (BtB)~\cite{bertazzini2024beyond}, SAGI-D~\cite{giakoumoglou2025sagi}, CocoGlide~\cite{guillaro2023trufor}, and B-Free~\cite{guillaro2025bias} reflect this shift toward realistic local edits.

Pretrained vision backbones have become central to image forensics. TruFor~\cite{guillaro2023trufor} leverages a pretrained vision transformer architecture to process RGB images together with learned noise-sensitive traces for general forgery localization. In parallel, self-supervised ViT backbones such as DINO~\cite{caron2021emerging} provide transferable dense representations that can be adapted to localize subtle generative traces. DinoLizer~\cite{butora2025dinolizer} demonstrates the value of DINOv2~\cite{oquab2023dinov2} patch embeddings for generative inpainting localization and reports strong cross-dataset performance.

However, existing DINO-based localization methods typically rely on final-layer features~\cite{butora2025dinolizer,yu2026dinov3}. Forensic methods primarily rely on low-level artifacts such as compression and sensor noise~\cite{mareen2024tgif}, while modern vision foundation models encode higher-level semantic structure in image representations, motivating the use of multi-level features. Multi-level fusion is a standard principle in dense prediction~\cite{lin2017feature,ronneberger2015u,chen2017rethinking}; for ViTs, intermediate hidden states provide a natural hierarchy of local, structural, and semantic information.

We propose \textbf{DINO-FIL}, a dense \textbf{F}orensic \textbf{I}npainting \textbf{L}ocalization framework built on a pretrained DINOv3 ViT-H+/16 encoder and a lightweight multi-layer fusion decoder. DINO-FIL extracts hidden states from several backbone depths, projects them to a common channel dimension, enriches them with lightweight multi-scale context modules inspired by atrous spatial pyramid pooling (ASPP)~\cite{chen2017rethinking}, and fuses them in a top-down decoder inspired by feature pyramid networks (FPN)~\cite{lin2017feature}. Since all selected DINOv3 hidden states share the same patch-grid resolution, DINO-FIL treats transformer depth, rather than spatial scale, as the fusion hierarchy. This enables the model to combine local appearance cues, texture- and pattern-sensitive intermediate representations, and higher-level semantic context. We further study partial encoder fine-tuning with a lower learning rate for unfrozen DINOv3 layers than for the newly initialized decoder.

In summary, our contributions are threefold:
\begin{itemize}
\item We introduce DINO-FIL, a DINOv3-based dense inpainting localization model that fuses intermediate transformer features from multiple backbone depths rather than relying only on the final layer.
\item We systematically study encoder adaptation and show that partial fine-tuning of later DINOv3 layers improves inpainting localization while full encoder fine-tuning provides limited additional benefit.
\item We provide a cross-dataset evaluation of our proposed approach against strong forensic baselines and show that high-resolution inference substantially improves inpainting localization, while maintaining low false-positive rates on pristine images.
\end{itemize}

\section{Related Work}
\label{sec:related_work}

\subsection{Pre-trained Vision Transformers in Image Forensics}
The adaptation of pre-trained Vision Transformers (ViTs) has emerged as a dominant paradigm for discovering localized anomalies in visual media. A prominent milestone in this space is TruFor \cite{guillaro2023trufor}, which utilizes a pre-trained SegFormer backbone architecture coupled with a learned noise-print stream to map localized forgery traces. While TruFor remains a heavily cited baseline, it is engineered for general image manipulation detection and localization (IMDL). Consequently, its training regimes and evaluation benchmarks encompass a wide variety of traditional manipulations (such as splicing, copy-move, and removal) rather than dedicating focus to the specific artifacts of generative AI inpainting.

Parallel to traditional end-to-end architectures, an emerging paradigm in dense forensic tasks explores leveraging the robust representation spaces of massive vision foundation models. DinoLizer \cite{butora2025dinolizer} establishes that self-supervised representations from DINOv2 contain rich, localized artifacts useful for dense binary forgery segmentation. To handle arbitrary image dimensions, overlapping image crops are cut out of the target image and all crops are processed by the DINO backbone sequentially. A 1$\times$1 convolutional layer is then used to infer localized logit maps directly from the patch embeddings, which are subsequently aggregated to construct the final heatmap.

Yu et al. \cite{yu2026dinov3} introduce a baseline that applies Low-Rank Adaptation (LoRA) to a frozen DINOv3 backbone paired with a lightweight convolutional decoder. Similar to TruFor, Yu et al.'s framework targets broad image forensics across diverse manipulation types rather than focusing exclusively on generative AI inpainting.

\subsection{Architectural Distinctions in DINO-FIL}
Our proposed framework, DINO-FIL, fundamentally departs from these existing DINO-based approaches in two key architectural and task-specific dimensions:

\begin{itemize}
    \item Feature Extraction Strategy: While both DinoLizer and Yu et al. rely exclusively on extracting features from the final layer of the DINO backbone, DINO-FIL systematically taps into the internal depth hierarchy of a high-capacity ViT-H+ encoder -- extracting intermediate hidden states from layers 8, 16, 24, and 32 to capture both low-level texture boundaries and high-level semantic inconsistencies.
    \item Fine-Tuning \& Learning Rate Paradigm: While both DinoLizer and Yu et al. evaluate full-parameter fine-tuning of the DINO encoder, their implementations default to training the entire backbone, utilizing a single global learning rate across both the encoder and decoder networks.
    In contrast, DINO-FIL adopts a more targeted optimization strategy. We investigate both full-parameter fine-tuning and fine-tuning restricted to the later transformer layers, and introduce discriminative learning rates that decouple the adaptation pace of the pretrained backbone from that of the newly initialized decoder. This helps preserve the integrity of the backbone’s foundational features while allowing the decoder to specialize for the localization task.
\end{itemize}

\section{Methodology}

\subsection{Model Architecture}


\begin{figure}[!t]
\centering
\resizebox{\linewidth}{!}{%
\begin{tikzpicture}[
    font=\small,
    >=Latex,
    block/.style={
        rectangle,
        draw=black!70,
        rounded corners=3pt,
        align=center,
        inner xsep=6pt,
        inner ysep=5pt,
        fill=white
    },
    inputblock/.style={
        block,
        text width=2.35cm,
        minimum height=0.95cm,
        fill=green!7
    },
    dinoblock/.style={
        block,
        text width=3.45cm,
        minimum width=3.70cm,
        minimum height=2.38cm,
        fill=blue!5
    },
    stage/.style={
        block,
        text width=3.05cm,
        minimum width=3.38cm,
        minimum height=1.58cm,
        fill=gray!6
    },
    projblock/.style={
        block,
        text width=3.20cm,
        minimum width=3.55cm,
        minimum height=1.48cm,
        fill=gray!6
    },
    contextblock/.style={
        block,
        text width=3.75cm,
        minimum width=4.05cm,
        minimum height=1.48cm,
        fill=gray!6
    },
    dropoutblock/.style={
        block,
        text width=3.75cm,
        minimum width=4.05cm,
        minimum height=0.86cm,
        fill=yellow!14
    },
    layer/.style={
        rectangle,
        draw=black!55,
        rounded corners=2pt,
        align=center,
        minimum width=1.28cm,
        minimum height=0.38cm,
        font=\scriptsize,
        fill=white
    },
    feat/.style={
        block,
        minimum width=1.05cm,
        minimum height=0.46cm,
        font=\scriptsize,
        fill=blue!5
    },
    phi/.style={
        block,
        minimum width=1.10cm,
        minimum height=0.58cm,
        font=\scriptsize,
        fill=orange!10
    },
    add/.style={
        circle,
        draw=black!70,
        minimum size=0.50cm,
        inner sep=0pt,
        font=\scriptsize,
        fill=orange!6
    },
    outblock/.style={
        block,
        minimum height=0.74cm,
        font=\scriptsize,
        fill=green!7
    },
    arrow/.style={-Latex, thick, draw=black!82},
    line/.style={thick, draw=black!82},
    bus/.style={thick, draw=black!60},
    lab/.style={font=\scriptsize, fill=white, inner sep=1.1pt},
    fusiontitle/.style={font=\footnotesize\bfseries, fill=white, inner sep=1.5pt}
]

\node[inputblock] (input) at (0.00,7.80)
    {Resized images\\batch\\[-1pt]
    $B\times3\times S\times S$};

\node[dinoblock] (dino) at (3.65,7.80) {};
\node[font=\small\bfseries] at ([yshift=0.78cm]dino.center) {DINOv3 encoder};
\node[font=\scriptsize] at ([yshift=0.44cm]dino.center) {selected hidden states};
\node[layer] (l8)  at ([xshift=-0.82cm,yshift=-0.03cm]dino.center) {Layer 8};
\node[layer] (l16) at ([xshift=0.82cm,yshift=-0.03cm]dino.center) {Layer 16};
\node[layer] (l24) at ([xshift=-0.82cm,yshift=-0.57cm]dino.center) {Layer 24};
\node[layer] (l32) at ([xshift=0.82cm,yshift=-0.57cm]dino.center) {Layer 32};
\node[font=\scriptsize] at ([yshift=-1.00cm]dino.center)
    {$h_\ell: B\times(q^2+5)\times C_b$};

\node[stage] (tok) at (8.10,7.80) {\textbf{Token-to-map}\\[0.5mm]
    \scriptsize drop CLS + 4 registers\\
    \scriptsize keep $q^2$ patch tokens\\
    \scriptsize reshape to $f_\ell$\\[0.4mm]
    $f_\ell: B\times C_b\times q\times q$};

\node[projblock] (proj) at (12.45,7.80)
    {\textbf{Projection adapter}\\[0.5mm]
    \scriptsize $1{\times}1$ adapter +\\
    GN + GELU\\[0.4mm]
    $p_\ell=A_\ell(f_\ell)$\\[0.2mm]
    $p_\ell: B\times C_d\times q\times q$};

\draw[arrow] (input.east) -- node[lab, above=5pt, pos=0.55] {$x$} (dino.west);
\draw[arrow] (dino.east) -- node[lab, above=5pt, pos=0.50] {$\{h_\ell\}$} (tok.west);
\draw[arrow] (tok.east) -- node[lab, above=5pt, pos=0.50] {$\{f_\ell\}$} (proj.west);

\node[contextblock] (ctx) at (12.45,5.45)
    {\textbf{Context block}\\[0.5mm]
    \scriptsize DW-sep $3{\times}3$ branches\\
    $d=1,2,4$; sum + $1{\times}1$ fuse\\[-0.1mm]
    $g_\ell=\psi_\ell(\sum_d D_\ell^{(d)}(p_\ell))$\\[0.2mm]
    $g_\ell: B\times C_d\times q\times q$};

\draw[arrow] (proj.south) -- node[lab, right=5pt, pos=0.50] {$\{p_\ell\}$} (ctx.north);

\node[dropoutblock] (drop) at (12.45,3.45)
    {\textbf{Optional Dropout2d}\\[-0.3mm]
     \scriptsize train only; identity otherwise\\[-0.2mm]
     $\tilde{g}_\ell=\operatorname{Dropout2d}(g_\ell;p)$};

\draw[arrow] (ctx.south) -- node[lab, right=5pt, pos=0.50] {$\{g_\ell\}$} (drop.north);

\coordinate (busL) at (1.05,2.35);
\coordinate (busR) at (11.75,2.35);
\coordinate (dropJoin) at (12.45,2.35);
\draw[arrow] (drop.south) -- node[lab, right=5pt, pos=0.50] {$\{\tilde{g}_\ell\}$} (dropJoin);
\draw[bus] (dropJoin) -- (busR);
\draw[bus] (busL) -- (busR);
\node[lab] at ($(busL)!0.50!(busR)+(0,0.25)$)
    {context maps $\{\tilde{g}_8,\tilde{g}_{16},\tilde{g}_{24},\tilde{g}_{32}\}$};

\node[phi] (phi32) at (1.05,0.15) {$\phi_{32}$};
\node[add] (add24) at (2.65,0.15) {$+$};
\node[phi] (phi24) at (4.25,0.15) {$\phi_{24}$};
\node[add] (add16) at (5.85,0.15) {$+$};
\node[phi] (phi16) at (7.45,0.15) {$\phi_{16}$};
\node[add] (add8) at (9.05,0.15) {$+$};
\node[phi] (phi8) at (10.65,0.15) {$\phi_8$};

\node[feat] (g32) at ($(phi32)+(0,1.20)$) {$\tilde{g}_{32}$};
\node[feat] (g24) at ($(add24)+(0,1.20)$) {$\tilde{g}_{24}$};
\node[feat] (g16) at ($(add16)+(0,1.20)$) {$\tilde{g}_{16}$};
\node[feat] (g8)  at ($(add8)+(0,1.20)$) {$\tilde{g}_8$};

\draw[arrow] (1.05,2.35) -- (g32.north);
\draw[arrow] (2.65,2.35) -- (g24.north);
\draw[arrow] (5.85,2.35) -- (g16.north);
\draw[arrow] (9.05,2.35) -- (g8.north);

\draw[arrow] (g32.south) -- (phi32.north);

\draw[arrow] (phi32.east) -- (add24.west);
\node[lab] at ($(phi32.east)!0.5!(add24.west)+(0,0.21)$) {$m_{32}$};
\draw[arrow] (g24.south) -- (add24.north);
\draw[arrow] (add24.east) -- (phi24.west);

\draw[arrow] (phi24.east) -- (add16.west);
\node[lab] at ($(phi24.east)!0.5!(add16.west)+(0,0.21)$) {$m_{24}$};
\draw[arrow] (g16.south) -- (add16.north);
\draw[arrow] (add16.east) -- (phi16.west);

\draw[arrow] (phi16.east) -- (add8.west);
\node[lab] at ($(phi16.east)!0.5!(add8.west)+(0,0.21)$) {$m_{16}$};
\draw[arrow] (g8.south) -- (add8.north);
\draw[arrow] (add8.east) -- (phi8.west);

\node[font=\scriptsize, align=center] (recur) at (5.70,-0.87)
    {$m_{32}=\phi_{32}(\tilde{g}_{32})$\quad
     $m_{24}=\phi_{24}(\tilde{g}_{24}+m_{32})$\quad
     $m_{16}=\phi_{16}(\tilde{g}_{16}+m_{24})$\quad
     $m_8=\phi_8(\tilde{g}_8+m_{16})$};

\node[fusiontitle] (flabel) at (5.70,-1.45)
    {Horizontal depth-wise top-down fusion, fixed $q\times q$ grid};

\node[outblock, text width=2.15cm, minimum width=2.48cm] (head) at (1.65,-3.08)
    {Prediction head\\[-1pt]
     $1{\times}1$ conv\\[-1pt]
     $B\times1\times q\times q$};
\node[outblock, text width=2.15cm, minimum width=2.52cm] (upsample) at (4.90,-3.08)
    {Bilinear\\[-1pt]
     upsampling\\[-1pt]
     $B\times1\times S\times S$};
\node[outblock, text width=1.38cm, minimum width=1.78cm] (sigmoid) at (7.75,-3.08)
    {Sigmoid};
\node[block, text width=2.95cm, minimum width=3.28cm, minimum height=0.86cm,
      font=\scriptsize, fill=green!9] (mask) at (10.82,-3.08)
    {Inpainting mask\\[-1pt]
    values in $[0,1]$\\[-1pt]
    $B\times1\times S\times S$};

\coordinate (m8Out) at ($(phi8.east)+(0.68,0)$);
\coordinate (routeLeft) at ($(head.north)+(0,0.33)$);
\coordinate (routeDown) at (m8Out|-routeLeft);
\draw[line] (phi8.east) -- (m8Out);
\node[lab] at ($(phi8.east)!0.5!(m8Out)+(0,0.24)$) {$m_8$};
\draw[arrow] (m8Out) -- (routeDown) -- (routeLeft) -- (head.north);
\draw[arrow] (head.east) -- (upsample.west);
\draw[arrow] (upsample.east) -- (sigmoid.west);
\draw[arrow] (sigmoid.east) -- (mask.west);

\end{tikzpicture}
}%
\caption{
Overview of DINO-FIL. The batch of resized images is encoded with DINOv3, and hidden states from layers $8, 16, 24,$ and $32$ are converted into spatial patch maps $f_\ell$ by reshaping and discarding one CLS token and four register tokens. The projection adapter maps $f_\ell$ to $p_\ell\in\mathbb{R}^{B\times C_d\times q\times q}$. The context block produces $g_\ell\in\mathbb{R}^{B\times C_d\times q\times q}$ using lightweight multi-scale depthwise-separable convolutions. Before fusion, $g_\ell$ is optionally regularized with spatial dropout, yielding $\tilde g_\ell$ during training and the identity mapping otherwise. The resulting $\tilde g_\ell$ maps are fused by a horizontal depth-wise top-down decoder on the fixed $q\times q$ patch grid. A $1{\times}1$ prediction head produces patch-level logits, which are upsampled and passed through a sigmoid to obtain the final inpainting probability mask. Here, $q=S/P$, $P=16$, $C_b=1280$, and $C_d=1024$.
}
\label{fig:dino_fil_architecture}
\end{figure}
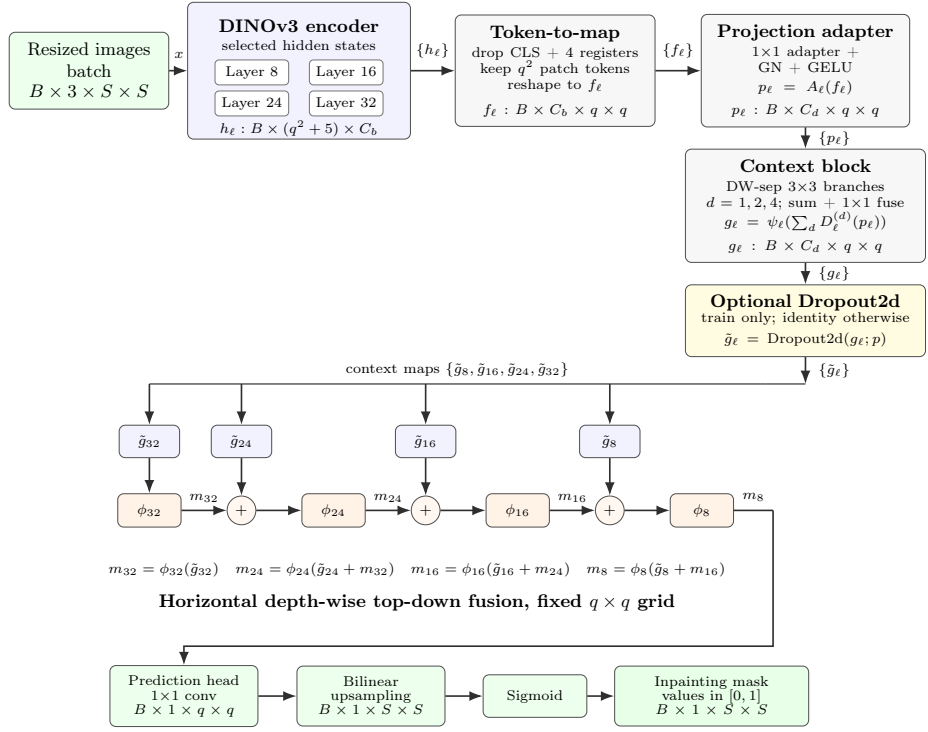

DINO-FIL is designed as a dense binary segmentation model for localizing inpainted image regions. As illustrated in \Cref{fig:dino_fil_architecture}, given a batch of RGB images \(x_0 \in \mathbb{R}^{B \times 3 \times H \times W}\), the images are resized to a square input resolution \(S \times S\), normalized, and passed through the network. The model outputs a single-channel logit map per image, which is converted with a sigmoid function into an inpainting probability mask \(\hat{M} \in [0,1]^{B \times 1 \times S \times S}\).

The encoder is a pretrained DINOv3 ViT-H+/16 backbone with patch size \(P=16\). Rather than using only the final transformer output, DINO-FIL extracts intermediate hidden states from layers \(8,16,24,\) and \(32\). These layers sample the 32-layer backbone at uniform depth intervals. This design is intended to expose the decoder to representations from different transformer depths: earlier hidden states can retain more local appearance and spatial detail, intermediate states can provide texture- and pattern-sensitive cues, and deeper states can contribute more semantic context useful for inpainting localization. The degree to which the DINOv3 backbone is fine-tuned is evaluated experimentally by unfreezing different numbers of final transformer layers, as described in  \Cref{sec:encoder_finetuning_ablation}.

For an input resolution \(S \times S\), the spatial patch grid has resolution \(q \times q\), where

\begin{equation}
q = \frac{S}{P}.
\label{eq:patch-grid-resolution}
\end{equation}
Each selected DINO hidden state contains \(N=q^2\) spatial patch tokens, followed by five non-spatial tokens: one CLS token and four register tokens. The non-spatial tokens are discarded before reshaping. Thus, for each selected layer \(\ell \in \{8,16,24,32\}\), the spatial tokens are converted into a patch-grid feature map
\begin{equation}
f_\ell \in \mathbb{R}^{B \times C_b \times q \times q},
\qquad
\ell \in \{8,16,24,32\}.
\label{eq:layer-feature-map}
\end{equation}
where \(C_b=1280\) is the DINO backbone channel dimension.

Each selected feature map is then passed through an independent projection adapter consisting of a \(1 \times 1\) convolution, group normalization, and GELU activation. The adapter maps the backbone channel dimension from \(C_b=1280\) to the decoder channel dimension \(C_d=1024\). This reduces downstream computational cost while preserving the shared \(q \times q\) patch-grid resolution.

To enrich the transformer features with local multi-scale spatial context, each projected feature map is processed by a lightweight ASPP-inspired context block. Let \(p_\ell\) denote the projected feature from layer \(\ell\). The context block applies three parallel depthwise separable \(3 \times 3\) convolution branches with dilation rates \(1\), \(2\), and \(4\). The branch outputs are summed and fused with a \(1 \times 1\) convolution:
\begin{equation}
g_\ell =
\psi_\ell\!\left(
D_\ell^{(1)}(p_\ell)
+
D_\ell^{(2)}(p_\ell)
+
D_\ell^{(4)}(p_\ell)
\right).
\label{eq:context-block}
\end{equation}
where \(D_\ell^{(d)}\) denotes a depthwise separable convolution with dilation \(d\), and \(\psi_\ell\) denotes the \(1 \times 1\) fusion convolution. This context module is useful for inpainting localization because manipulation traces may appear at different spatial scales, ranging from fine boundary discontinuities and local texture artifacts to broader semantic or blending inconsistencies.

The resulting context-enhanced features are optionally regularized with spatial dropout before fusion: 
\begin{equation}
\tilde{g}_\ell =
\begin{cases}
\operatorname{Dropout2d}(g_\ell; p), & p > 0 \text{ during training},\\
g_\ell, & \text{otherwise}.
\end{cases}
\label{eq:spatial-dropout}
\end{equation}
where \(p\) is the decoder dropout rate. The regularized features are then merged by an FPN-inspired depth-wise top-down fusion decoder. Unlike a conventional Feature Pyramid Network, the selected DINO hidden states do not form a spatial pyramid: all selected layers share the same \(q \times q\) patch-grid resolution, so no resizing is required between fusion stages. Instead, the decoder uses transformer depth as the fusion hierarchy, merging the selected hidden states from deepest to shallowest rather than across spatial scales.

Let \(\phi_\ell\) denote the learned post-fusion \(3 \times 3\) convolution for layer \(\ell\). Starting from the deepest selected feature, the decoder progressively adds shallower features and refines each merged representation:
\begin{equation}
\begin{aligned}
m_{32} &= \phi_{32}(\tilde{g}_{32}),\\
m_{24} &= \phi_{24}(\tilde{g}_{24}+m_{32}),\\
m_{16} &= \phi_{16}(\tilde{g}_{16}+m_{24}),\\
m_{8}  &= \phi_{8}(\tilde{g}_{8}+m_{16}).
\end{aligned}
\label{eq:depth-wise-fusion}
\end{equation}
This preserves the top-down fusion principle of FPN while adapting it to a transformer backbone with equal-resolution intermediate states.

Finally, a \(1 \times 1\) prediction head maps \(m_8\) to a single-channel patch-level logit map
\begin{equation}
z \in \mathbb{R}^{B \times 1 \times q \times q}.
\label{eq:patch-level-logits}
\end{equation}
The logits are bilinearly upsampled to the resized input resolution \(S \times S\) and passed through a sigmoid function:
\begin{equation}
\hat{M}
=
\sigma\!\left(\operatorname{Upsample}(z)\right),
\qquad
\hat{M} \in [0,1]^{B \times 1 \times S \times S}.
\label{eq:predicted-mask}
\end{equation}

The predicted mask is supervised with the binary inpainting ground-truth mask using the segmentation loss described in \Cref{sec:training_protocol}.

\section{Experimental Setup}
\label{sec:experimental_setup}

\subsection{Datasets}
We evaluate DINO-FIL on a cross-dataset image inpainting localization benchmark. To make the results comparable to DinoLizer, which uses the same dataset-level training setup, DINO-FIL is trained on B-Free and evaluated on Beyond the Brush, CocoGlide, TGIF, SAGI-D, and B-Free. The reorganized B-Free split contains 206,065 training images, 25,755 validation images, and 25,765 test images. However, for evaluation we follow the benchmark protocol and test on 10,000 B-Free test images. The B-Free test split is included to measure in-domain performance, while the remaining datasets are used to assess cross-dataset generalization. In total, DINO-FIL is tested on 28,835 images, comprising 10,000 B-Free, 3,896 Beyond the Brush, 512 CocoGlide, 4,249 TGIF, and 10,178 SAGI-D test images.

\subsection{Training Protocol}
\label{sec:training_protocol}
DINO-FIL is trained as a binary segmentation model for dense inpainting localization. The network predicts a single-channel logit map, which is supervised with the binary ground-truth manipulation mask. The training objective combines binary cross-entropy with logits and Dice loss:
\begin{equation}
\mathcal{L}
=
\mathcal{L}_{\mathrm{BCE}}
+
\mathcal{L}_{\mathrm{Dice}}.
\label{eq:training-loss}
\end{equation}
This objective encourages both accurate pixel-wise classification and spatial overlap between the predicted and ground-truth manipulation masks.

Unless otherwise stated, all models are trained using an input resolution of $512 \times 512$. Images are converted to RGB, resized to the target resolution, transformed into tensors, and normalized with ImageNet statistics. During training, we apply random horizontal flipping, color jitter, and B-Free-style image perturbations, including blur, JPEG recompression, down/up-scaling, and mild noise. At test time, only resizing, tensor conversion, and ImageNet normalization are applied.

The decoder is optimized with a learning rate of $1 \times 10^{-4}$, while unfrozen encoder layers are optimized with a lower learning rate of $5 \times 10^{-6}$. The batch size is set to 12, and the decoder dropout rate is set to 0.05. For all experiments, predicted probability maps are thresholded at 0.5 to obtain binary localization masks.

\subsection{Encoder Fine-Tuning Ablation}
\label{sec:encoder_finetuning_ablation}

DinoLizer reports that keeping the DINO backbone frozen can improve localization performance. To investigate whether this observation also holds for the proposed DINO-FIL architecture, we conduct an encoder fine-tuning ablation. Specifically, we train variants in which different numbers of final DINOv3 transformer layers are unfrozen while the remaining encoder layers stay frozen.

We evaluate six settings: 0, 6, 12, 16, 24, and 32 unfrozen transformer layers. The setting with 0 unfrozen layers corresponds to a fully frozen DINOv3 encoder, while the setting with 32 unfrozen layers fine-tunes the complete transformer backbone. All unfrozen encoder layers use the same encoder learning rate of $5 \times 10^{-6}$. This experiment tests whether adapting the pretrained DINOv3 backbone features, either partially or end-to-end, improves inpainting localization when used with the proposed decoder. Due to the large number of B-Free training images, all models in this experiment are trained for two epochs.

\subsection{Test Resolution Evaluation}
Although DINO-FIL is trained at $512 \times 512$, we evaluate the trained models at both $512 \times 512$ and $1024 \times 1024$ test resolutions. This experiment measures whether higher-resolution inference improves localization by preserving finer manipulation traces, boundary artifacts, and texture inconsistencies that may be suppressed at lower resolution.

The comparison is particularly relevant because the benchmark datasets differ in their native image resolutions. B-Free consists of fixed $512 \times 512$ images, whereas other datasets contain images with more diverse resolutions and aspect ratios. Consequently, evaluating only at $512 \times 512$ may underestimate performance on datasets where additional high-frequency evidence is available in the original image.

\subsection{Training Resolution Evaluation}
To further study the role of image resolution, we conduct a separate training-resolution experiment using SAGI-D as the training dataset. SAGI-D is suitable for this analysis because it contains images with a broad range of resolutions, including images smaller than 512 px and images larger than 1024 px. This contrasts with B-Free and CocoGlide, whose images have fixed resolutions of $512 \times 512$ and $256 \times 256$, respectively.

In this experiment, we compare models trained at $512 \times 512$ and $1024 \times 1024$. The $(512 \times 512)$-trained model is evaluated at both $512 \times 512$ and $1024 \times 1024$, while the $(1024 \times 1024)$-trained model is evaluated at $1024 \times 1024$. B-Free and CocoGlide are excluded from this experiment because their fixed low resolutions do not provide additional image information at $1024 \times 1024$; upsampling would merely duplicate existing pixels rather than introducing meaningful high-resolution evidence. Due to the large number of SAGI-D training images, all models in this experiment are trained for two epochs.

\subsection{Evaluation Metrics}
To enable direct comparison with DinoLizer, we follow the same pixel-level evaluation protocol for manipulated images and report Intersection over Union (IoU) and F1 score. The predicted probability map is binarized using a fixed threshold of 0.5. Let $TP$, $FP$, and $FN$ denote the number of true-positive, false-positive, and false-negative pixels, respectively. IoU is computed as shown in \cref{eq:iou}.
\begin{equation}
\mathrm{IoU}
=
\frac{TP}{TP + FP + FN}.
\label{eq:iou}
\end{equation}
Following DinoLizer, F1 is computed from precision and recall with a small stabilization constant $\epsilon=10^{-8}$:
\begin{equation}
\mathrm{Precision}
=
\frac{TP}{TP + FP + \epsilon},
\qquad
\mathrm{Recall}
=
\frac{TP}{TP + FN + \epsilon}.
\label{eq:precision-recall}
\end{equation}

\begin{equation}
\mathrm{F1}
=
2 \cdot
\frac{
\mathrm{Precision} \cdot \mathrm{Recall}
}{
\mathrm{Precision} + \mathrm{Recall} + \epsilon
}.
\label{eq:f1}
\end{equation}

Both IoU and F1 are computed independently for each manipulated image and then macro-averaged over all manipulated test images. Thus, each image contributes equally to the final score, independent of the size of the manipulated region. Since IoU and F1 are averaged separately after being computed per image, the final mean scores do not necessarily preserve the one-to-one monotonic relationship that holds for a single fixed set of $TP$, $FP$, and $FN$ values. Consequently, it is possible for one model to obtain a slightly lower mean IoU but a higher mean F1 than another model.

In addition to manipulated-image localization, we evaluate false positives on pristine images. For this analysis, we use original, non-manipulated images from the SAGI-D test split, consisting of COCO, Open Images, and RAISE images. We report the pixel-level false-positive rate and the image-level true-negative rate. An image is counted as a true negative if no pixel exceeds the decision threshold of 0.5.

\subsection{Comparison with Existing Detectors}
We compare DINO-FIL against DinoLizer and TruFor, the two strongest-performing methods among the six detectors compared in the DinoLizer benchmark. For both baselines, we use the results reported in the DinoLizer paper. The reported baseline results are not based on externally downsampling all test images to a fixed resolution before inference: DinoLizer avoids downsampling through sliding-window inference, and TruFor processes arbitrary-size inputs and produces localization maps at input resolution. Since DinoLizer reports separate results for SAGI-SP and SAGI-FR, whereas we evaluate SAGI-D as a unified dataset, we aggregate the reported SAGI-SP and SAGI-FR scores into a single SAGI-D score using an image-count-weighted average.\footnote{We aggregate the two subsets rather than interpreting SAGI-SP and SAGI-FR separately because the SP/FR assignment of individual generators is not fully consistent across sources. In particular, DinoLizer reports Inpaint-Anything under SAGI-FR, whereas the SAGI-D definition categorizes Inpaint-Anything as SP because it preserves the unmasked region.}

\section{Results}
\label{sec:results}

\subsection{Encoder Fine-Tuning and Test Resolution}

\begin{table*}[t]
\centering
\scriptsize
\setlength{\tabcolsep}{4pt}
\renewcommand{\arraystretch}{1.12}

\caption{Comparison of DINO-FIL variants under different numbers of unfrozen encoder layers and test image resolutions (training resolution is fixed at 512px), evaluated using average pixel-level IoU and F1 score on image forgery localization. The best results are shown in bold, and the second-best results are underlined.}
\label{tab:finetuning_test_resolution_results}
\resizebox{\textwidth}{!}{%
\begin{tabular}{llcccccccccccc}
\toprule
\multirow{2}{*}{\makecell{Test\\image\\resolution}}
& \multirow{2}{*}{\makecell{Unfrozen\\layers}}
& \multicolumn{2}{c}{BtB}
& \multicolumn{2}{c}{CocoGlide}
& \multicolumn{2}{c}{TGIF}
& \multicolumn{2}{c}{SAGI-D}
& \multicolumn{2}{c}{B-Free}
& \multicolumn{2}{c}{Average} \\
\cmidrule(lr){3-4}
\cmidrule(lr){5-6}
\cmidrule(lr){7-8}
\cmidrule(lr){9-10}
\cmidrule(lr){11-12}
\cmidrule(lr){13-14}
& & IoU & F1
& IoU & F1
& IoU & F1
& IoU & F1
& IoU & F1
& IoU & F1 \\
\midrule

\multirow{6}{*}{512px}
& 0 
& 0.23 & 0.31
& 0.75 & 0.82
& 0.27 & 0.33
& 0.47 & 0.55
& 0.82 & 0.89
& 0.51 & 0.58 \\

& 6 
& 0.33 & 0.41
& 0.77 & 0.85
& 0.38 & 0.45
& 0.52 & 0.61
& 0.86 & 0.92
& 0.57 & 0.65 \\

& 12 
& 0.36 & 0.44
& 0.81 & 0.87
& 0.38 & 0.44
& 0.58 & 0.66
& 0.89 & 0.94
& 0.60 & 0.67 \\

& 16 
& 0.36 & 0.43
& 0.83 & 0.88
& 0.38 & 0.43
& 0.59 & 0.67
& 0.90 & 0.94
& 0.61 & 0.67 \\

& 24 
& 0.32 & 0.38
& 0.84 & 0.89
& 0.33 & 0.37
& 0.57 & 0.64
& \underline{0.91} & \underline{0.95}
& 0.60 & 0.65 \\

& 32 
& 0.33 & 0.40
& 0.87 & 0.92
& 0.36 & 0.40
& 0.58 & 0.64
& \textbf{0.91} & \textbf{0.95}
& 0.61 & 0.66 \\

\midrule

\multirow{6}{*}{1024px}
& 0 
& 0.38 & 0.50
& 0.75 & 0.83
& 0.42 & 0.51
& 0.53 & 0.64
& 0.80 & 0.88
& 0.58 & 0.67 \\

& 6 
& 0.47 & 0.57
& 0.77 & 0.85
& 0.47 & 0.56
& 0.57 & 0.66
& 0.83 & 0.90
& 0.62 & 0.71 \\

& 12 
& 0.54 & 0.64
& 0.84 & 0.90
& 0.52 & 0.59
& 0.63 & 0.71
& 0.88 & 0.93
& 0.68 & 0.75 \\

& 16 
& 0.56 & 0.65
& 0.85 & 0.91
& 0.54 & 0.61
& \textbf{0.66} & \textbf{0.73}
& 0.89 & 0.94
& 0.70 & 0.77 \\

& 24 
& \textbf{0.59} & \textbf{0.68}
& \underline{0.87} & \underline{0.92}
& \underline{0.57} & \underline{0.63}
& \underline{0.66} & \underline{0.73}
& 0.87 & 0.92
& \textbf{0.71} & \textbf{0.77} \\

& 32 
& \underline{0.57} & \underline{0.66}
& \textbf{0.88} & \textbf{0.93}
& \textbf{0.58} & \textbf{0.64}
& 0.65 & 0.72
& 0.88 & 0.92
& \underline{0.71} & \underline{0.77} \\

\bottomrule
\end{tabular}%
}

\end{table*}

\Cref{tab:finetuning_test_resolution_results} reports the localization performance of DINO-FIL when different numbers of final DINOv3 transformer layers are unfrozen and the model is evaluated at different test image resolutions. The results show that adapting the encoder yields higher test performance for the proposed architecture. With a fully frozen encoder, DINO-FIL achieves an average IoU/F1 of 0.51/0.58 at a test resolution of $512 \times 512$ and 0.58/0.67 at $1024 \times 1024$. At $1024 \times 1024$, unfreezing later transformer blocks consistently improves performance up to 24 unfrozen layers, although the gains become smaller as more layers are unfrozen. At 1024 × 1024 resolution, the model reaches its peak average IoU/F1 with 24 unfrozen layers. Fully unfreezing all 32 layers does not further improve the average result and slightly decreases the average F1 before rounding, from 0.774 to 0.771. This suggests that partial fine-tuning of the upper DINOv3 layers is sufficient, while full encoder fine-tuning provides diminishing returns.

A second clear trend is the effect of test resolution. Evaluating at $1024 \times 1024$ substantially improves performance on the cross-dataset test sets. Comparing the best-performing configurations at each resolution, the average IoU/F1 increases from 0.61/0.67 at $512 \times 512$ to 0.71/0.77 at $1024 \times 1024$. This improvement is most pronounced on datasets with larger or more diverse native image resolutions: Beyond the Brush increases from 0.36/0.44 to 0.59/0.68, TGIF from 0.38/0.45 to 0.58/0.64, and SAGI-D from 0.59/0.67 to 0.66/0.73. These results suggest that higher-resolution inference can preserve manipulation traces, boundary inconsistencies, and local texture artifacts that may be suppressed when images are resized to $512 \times 512$, making $1024 \times 1024$ evaluation better aligned with the available image evidence for higher-resolution datasets.

CocoGlide also shows a small improvement, increasing from 0.87/0.92 to 0.88/0.93. However, this gain should be interpreted cautiously because CocoGlide images have a fixed resolution of $256 \times 256$ pixels. Evaluating these images at $1024 \times 1024$ therefore does not introduce additional high-frequency information, but only upsamples the existing pixels. The small observed increase may consequently be due to interpolation effects or random variation rather than a genuine resolution benefit.

The only exception is B-Free, where $512 \times 512$ evaluation performs slightly better than $1024 \times 1024$ evaluation. This is expected because B-Free consists of fixed $512 \times 512$ images and is also the training dataset for this experiment. As with CocoGlide, testing B-Free at $1024 \times 1024$ does not introduce additional image information, but instead upsamples the same image content.

Overall, the best average performance is obtained by the model with 24 unfrozen layers evaluated at $1024 \times 1024$, achieving an average IoU/F1 of 0.71/0.77, with the fully fine-tuned model with 32 unfrozen layers obtaining nearly identical average performance. This indicates that partial fine-tuning of the final DINOv3 layers is sufficient for strong localization performance, while full end-to-end fine-tuning does not provide a clear additional benefit.

\subsection{Comparison with Existing Detectors}

\begin{table*}[t]
\centering
\caption{Comparison of DINO-FIL (ours) with DinoLizer and TruFor, the two strongest-performing methods reported in the DinoLizer benchmark~\cite{butora2025dinolizer}. Performance is reported using average pixel-level IoU and F1 score for image forgery localization. The SAGI-D results for DinoLizer and TruFor are computed as image-count-weighted averages of the reported SAGI-SP and SAGI-FR results. The best results are shown in bold, and the second-best results are underlined.}
\label{tab:prior_work_comparison}

\resizebox{\textwidth}{!}{%
\begin{tabular}{l|cc|cc|cc|cc|cc|cc}
\hline
\multirow{2}{*}{Model}
& \multicolumn{2}{c|}{BtB}
& \multicolumn{2}{c|}{CocoGlide}
& \multicolumn{2}{c|}{TGIF}
& \multicolumn{2}{c|}{SAGI-D}
& \multicolumn{2}{c|}{B-Free}
& \multicolumn{2}{c}{Average} \\
& IoU & F1
& IoU & F1
& IoU & F1
& IoU & F1
& IoU & F1
& IoU & F1 \\
\hline

\textbf{DINO-FIL (ours)}
& \textbf{0.59} & \textbf{0.68}
& \textbf{0.87} & \textbf{0.92}
& \underline{0.57} & \underline{0.63}
& \textbf{0.66} & \textbf{0.73}
& \textbf{0.87} & \textbf{0.92}
& \textbf{0.71} & \textbf{0.77} \\

DinoLizer
& \underline{0.42} & \underline{0.56}
& \underline{0.57} & \underline{0.69}
& 0.39 & 0.49
& \underline{0.38} & \underline{0.49}
& \underline{0.70} & \underline{0.81}
& \underline{0.49} & \underline{0.61} \\

TruFor
& 0.22 & 0.30
& 0.29 & 0.36
& \textbf{0.70} & \textbf{0.80}
& 0.35 & 0.41
& 0.22 & 0.30
& 0.36 & 0.43 \\

\hline
\end{tabular}%
}

\end{table*}

\Cref{tab:prior_work_comparison} compares DINO-FIL with DinoLizer and TruFor. For this comparison, we use the strongest-performing DINO-FIL variant from \Cref{tab:finetuning_test_resolution_results}, which unfreezes the final 24 DINOv3 transformer layers and is evaluated at $1024 \times 1024$ resolution. DINO-FIL achieves the best average performance across the benchmark, with an average IoU/F1 of 0.71/0.77. This is substantially higher than DinoLizer, which obtains 0.49/0.61, and TruFor, which obtains 0.36/0.43. DINO-FIL also achieves the best performance on four of the five evaluated datasets: Beyond the Brush, CocoGlide, SAGI-D, and B-Free.

The largest gains over DinoLizer are observed on CocoGlide and SAGI-D. On CocoGlide, DINO-FIL improves IoU from 0.57 to 0.87 and F1 from 0.69 to 0.92. On SAGI-D, DINO-FIL improves IoU from 0.38 to 0.66 and F1 from 0.49 to 0.73. These results indicate that the proposed multi-layer DINOv3 feature fusion decoder generalizes well across inpainting datasets and benefits from high-resolution inference.

TruFor achieves the strongest result on TGIF, with an IoU/F1 of 0.70/0.80, while DINO-FIL obtains 0.57/0.63. This is the only dataset on which DINO-FIL does not obtain the best result. Nevertheless, DINO-FIL remains the second-best method on TGIF and outperforms TruFor by a large margin on all other datasets.

The comparison should be interpreted as a benchmark-level comparison rather than a fully controlled retraining study. The DinoLizer and TruFor scores are taken from the DinoLizer benchmark. TruFor, however, was not trained on TGIF, as stated in the TruFor paper. Despite this limitation, the results show that DINO-FIL provides strong cross-dataset localization performance relative to the best-performing methods reported in the DinoLizer benchmark.

\subsection{False Positives on Pristine Images}

\begin{table*}[t]
\centering
\caption{False-positive evaluation of DINO-FIL on original, non-manipulated images. We report the pixel-level false-positive rate and the image-level true-negative rate. The average is computed over all images across the three datasets.}
\label{tab:original_fp_tn_results}

\begin{tabular*}{\textwidth}{@{\extracolsep{\fill}}l c cc}
\hline
Dataset
& Number of Images
& Pixel-level FP Rate
& Image-level TN Rate \\
\hline

COCO
& 2,922
& 0.0019
& 0.9767 \\

Open Images
& 5,579
& 0.0040
& 0.9684 \\

RAISE
& 580
& 0.0021
& 0.9810 \\

\hline
Average
& 9,081
& 0.0032
& 0.9719 \\

\hline
\end{tabular*}

\end{table*}

\Cref{tab:original_fp_tn_results} evaluates DINO-FIL on pristine, non-manipulated images from the SAGI-D test split. For this evaluation, we use the strongest-performing DINO-FIL variant from \Cref{tab:finetuning_test_resolution_results}, which unfreezes the final 24 DINOv3 transformer layers and is evaluated at $1024 \times 1024$ resolution. Across 9,081 original images from COCO, Open Images, and RAISE, the model obtains a pixel-level false-positive rate of 0.0032 and an image-level true-negative rate of 0.9719 at the fixed threshold of 0.5.

DINO-FIL achieves high true-negative rates across all three pristine-image subsets, with 97.67\% on COCO, 96.84\% on Open Images, and 98.10\% on RAISE. These results suggest that DINO-FIL is not only effective at localizing manipulated regions but also relatively conservative on pristine images. However, a low false-positive rate does not by itself guarantee that every predicted region in a manipulated image is spatially correct. Rather, it indicates that, under the chosen threshold, predicted inpainted regions are uncommon on pristine-image inputs.

\subsection{Effect of Training Resolution}

\begin{table*}[t]
\centering
\caption{Comparison of DINO-FIL variants trained at different resolutions (512px and 1024px) and evaluated at corresponding test resolutions for image forgery localization. Performance is reported in terms of average pixel-level IoU and F1 score across datasets. The best results are shown in bold, and the second-best results are underlined.}
\label{tab:training_resolution_results}

\scriptsize
\setlength{\tabcolsep}{4pt}
\renewcommand{\arraystretch}{1.12}

\resizebox{\textwidth}{!}{%
\begin{tabular}{llcccccccccccc}
\toprule
\multirow{2}{*}{\makecell{Train\\Resolution}}
& \multirow{2}{*}{\makecell{Test\\Resolution}}
& \multicolumn{2}{c}{Beyond the Brush}
& \multicolumn{2}{c}{TGIF}
& \multicolumn{2}{c}{SAGI-D}
& \multicolumn{2}{c}{Average} \\
\cmidrule(lr){3-4}
\cmidrule(lr){5-6}
\cmidrule(lr){7-8}
\cmidrule(lr){9-10}
& & IoU & F1
& IoU & F1
& IoU & F1
& IoU & F1 \\
\midrule

\multirow{2}{*}{512px}
& 512px
& 0.37 & 0.44
& 0.41 & 0.44
& 0.76 & 0.81
& 0.51 & 0.56 \\

& 1024px
& \textbf{0.68} & \textbf{0.76}
& \textbf{0.62} & \textbf{0.69}
& \underline{0.83} & \textbf{0.88}
& \textbf{0.71} & \textbf{0.78} \\

\midrule

1024px
& 1024px
& \underline{0.64} & \underline{0.71}
& \underline{0.55} & \underline{0.62}
& \textbf{0.84} & \textbf{0.88}
& \underline{0.68} & \underline{0.74} \\

\bottomrule
\end{tabular}%
}

\end{table*}

Motivated by the superior performance observed when evaluating DINO-FIL at $1024 \times 1024$, \Cref{tab:training_resolution_results} studies whether this benefit also transfers to training at higher resolution. All models in this experiment are based on the strongest-performing encoder configuration identified in \Cref{tab:finetuning_test_resolution_results}, i.e., DINO-FIL with the final 24 DINOv3 transformer layers unfrozen.

This experiment is independent from the B-Free training experiment because the models are trained on SAGI-D instead. Therefore, the absolute numbers should not be compared directly with the B-Free-trained models in \Cref{tab:finetuning_test_resolution_results}.

The results show that high-resolution inference also improves performance for DINO-FIL trained on SAGI-D at $512 \times 512$. For the model trained at this resolution, average IoU/F1 increases from 0.51/0.56 when evaluated at $512 \times 512$ to 0.71/0.78 when evaluated at $1024 \times 1024$.

However, training directly at $1024 \times 1024$ does not further improve average performance in this experiment. The model trained and tested at $1024 \times 1024$ achieves an average IoU/F1 of 0.68/0.74, which is below the $512 \times 512$-trained model evaluated at $1024 \times 1024$. The only dataset where $1024 \times 1024$ training slightly improves IoU is SAGI-D, where IoU increases from 0.83 to 0.84 while F1 remains unchanged at 0.88.

Several factors may explain this result. First, most SAGI-D images are not substantially larger than $1024$ px: only about 21\% have a maximum side length of at least $1024$ px, while the majority fall between $512$ px and $1024$ px. Consequently, training at $1024 \times 1024$ may not consistently expose the model to genuinely higher-resolution evidence. 

Second, training at $512 \times 512$ may act as a form of scale regularization. When combined with $1024 \times 1024$ inference, the model may still exploit high-frequency cues in the test images while retaining stronger cross-dataset generalization from the lower-resolution training setting.

These findings suggest that high-resolution inference is beneficial for DINO-FIL, whereas high-resolution training does not provide an additional benefit under the fixed-resolution setup evaluated here. This indicates that the relationship between training resolution, test resolution, and dataset image statistics requires further investigation.

\section{Conclusion}
\label{sec:conclusion}

This paper presented DINO-FIL, a dense inpainting localization model that combines intermediate DINOv3 representations with a lightweight multi-layer feature fusion decoder. Instead of relying only on the final transformer output, DINO-FIL exploits hidden states from multiple backbone depths and fuses them through projection, context aggregation, and top-down refinement. This design allows the model to combine local appearance cues, texture-sensitive intermediate representations, and deeper semantic information for pixel-level inpainting localization.

The experimental results show that DINO-FIL achieves strong cross-dataset generalization when trained on B-Free and evaluated on five inpainting localization benchmarks. Partial fine-tuning of the DINOv3 encoder improves performance over a fully frozen backbone, with the best average result obtained by unfreezing the final 24 transformer layers and evaluating at $1024 \times 1024$ resolution. Compared with DinoLizer and TruFor, DINO-FIL achieves the highest average IoU and F1 score and obtains the best performance on four of the five evaluated datasets. The false-positive analysis on pristine SAGI-D images further shows that the model remains conservative on non-manipulated inputs, achieving a low pixel-level false-positive rate and a high image-level true-negative rate.

The resolution experiments indicate that high-resolution inference is particularly important for inpainting localization. Evaluating at $1024 \times 1024$ substantially improves performance on datasets with larger or more diverse native image resolutions, suggesting that fine manipulation traces, boundary artifacts, and texture inconsistencies can be suppressed by lower-resolution resizing. However, training directly at $1024 \times 1024$ does not provide a consistent additional benefit in the fixed-resolution setup evaluated here. Instead, it yields slightly lower average performance than training at $512 \times 512$ and evaluating at $1024 \times 1024$. This suggests that the relationship between training resolution, test resolution, and dataset image statistics is more nuanced than simply increasing both resolutions.

\subsection{Future Work}

The observation that high-resolution inference is beneficial, whereas high-resolution training does not provide an additional benefit under the fixed-resolution setup evaluated here, suggests that the interaction between training resolution, test resolution, and dataset image statistics requires further investigation.

Future work should therefore investigate adaptive-resolution training and evaluation strategies. One possible approach is to group images according to their native resolution and process them at higher target resolutions only when sufficient high-resolution evidence is available. Since the DINOv3 backbone uses a patch size of 16, these target resolutions should be chosen as multiples of 16 to avoid implicit cropping of the input image as described in the DINOv3 model card~\cite{dinov3_modelcard}.

Another promising direction is parameter-efficient adaptation of the backbone using Low-Rank Adaptation (LoRA). Compared to full or partial fine-tuning of transformer layers, LoRA introduces trainable low-rank updates while keeping the pretrained weights frozen, potentially improving efficiency and reducing overfitting. Yu et al.~\cite{yu2026dinov3} explored LoRA-style adaptation for DINO-based forensic backbones, suggesting that it may improve performance compared to full fine-tuning. Integrating LoRA with the proposed multi-layer feature fusion architecture is therefore a promising direction for enhancing the DINO’s representational capacity.

\subsection*{Acknowledgements}
This work was supported by the GADMO 2.0 project, co-funded by the European Union’s Digital Europe Programme (DIGITAL) and the Austrian Research Promotion Agency (FFG), under grant agreement No. 101083573.

%
%

\end{document}